\documentclass{article}

\usepackage[final,nonatbib]{nips_2016}

\usepackage[utf8]{inputenc} 
\usepackage[T1]{fontenc}    
\usepackage{hyperref}       
\usepackage{url}            
\usepackage{booktabs}       
\usepackage{amsfonts}       
\usepackage{nicefrac}       
\usepackage{microtype}      

\usepackage{amssymb,amsmath,epsfig}
\usepackage{graphicx}
\usepackage{times} 
\usepackage{color}
\usepackage{wrapfig}
\usepackage{epsf}
\usepackage{colortbl}
\usepackage{verbatim}
\usepackage{subfigure}
\usepackage{multirow,rotating}
\usepackage{fancyheadings} 
\usepackage{xcolor} 
\usepackage{color, colortbl}
\usepackage{amstext}
\usepackage{paralist}
\usepackage{enumitem}
\usepackage{pdfpages}

\usepackage{algorithm}
\usepackage{algpseudocode}
\usepackage{amssymb}
\usepackage{amsmath}
\usepackage{lscape}

\title{Training an Interactive Humanoid Robot Using Multimodal Deep Reinforcement Learning}

\author{
  Heriberto Cuay\'ahuitl$^{1}$ and Guillaume Couly$^{2}$ and Cl\'ement Olalainty$^{2}$  \\
  $^{1}$University of Lincoln, School of Computer Science, United Kingdom\\
  $^{2}$Robotics and Interactive Systems, University of Toulouse III, Paul Sabatier, France \\
  \texttt{HCuayahuitl@lincoln.ac.uk} \\
}

\begin{document}

\maketitle

\begin{abstract}
Training robots to perceive, act and communicate using multiple modalities still represents a challenging problem, particularly if robots are expected to learn efficiently from small sets of example interactions. We describe a learning approach as a step in this direction, where we teach a humanoid robot how to play the game of noughts and crosses. Given that multiple multimodal skills can be trained to play this game, we focus our attention to training the robot to perceive the game, and to interact in this game. Our multimodal deep reinforcement learning agent perceives multimodal features and exhibits verbal and non-verbal actions while playing. Experimental results using simulations show that the robot can learn to win or draw up to 98\% of the games. A pilot test of the proposed multimodal system for the targeted game—integrating speech, vision and gestures---reports that reasonable and fluent interactions can be achieved using the proposed approach.
\end{abstract}


\section{Introduction}
\label{Intro}
\begin{figure*}
\begin{flushleft}
\centerline{\hspace{0.15cm} 
    \includegraphics[width=0.45\textwidth]{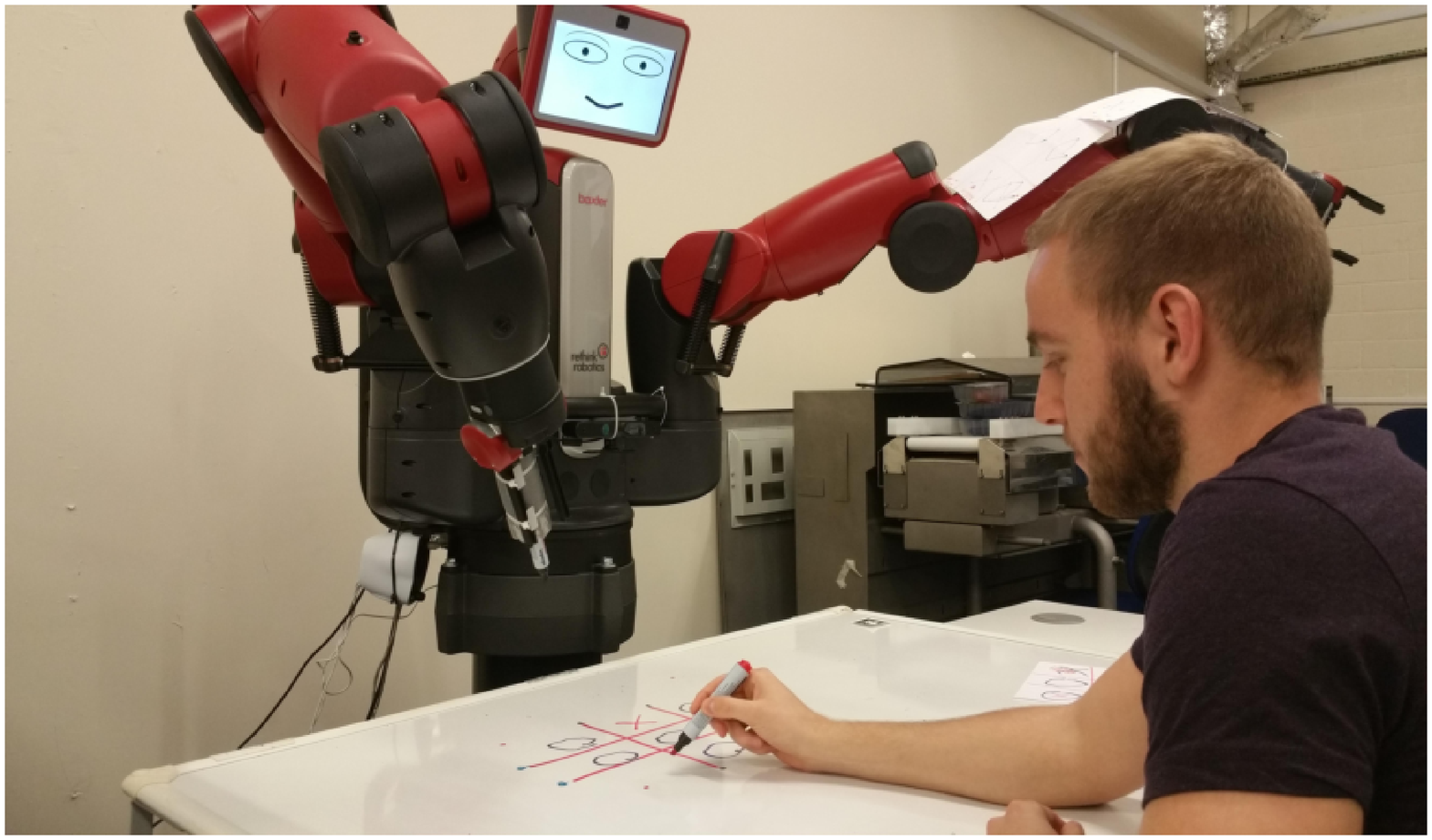}
\hspace{-0.15cm}
\scriptsize
\begin{tabular}{l}
Rob: Hello! \\
Rob: I am Baxter. \\
Rob: Would you like to play a game with me? \\
Usr: Yes, let's go for it.\\
Rob: Nice. Let me start.\\
Rob: I take this one $[who$=$rob \wedge what$=$draw \wedge where$=$lowermiddle]$\\
Rob: Your turn.\\
Usr: I pick this $[who$=$usr \wedge what$=$draw \wedge where$=$middleleft]$\\    
Rob: I take this one $[who$=$rob \wedge what$=$draw \wedge where$=$lowerright]$\\
Rob: Your turn.\\
Usr: I do this $[who$=$usr \wedge what$=$draw \wedge where$=$middle]$\\
Rob: I take this one $[who$=$rob \wedge what$=$draw \wedge where$=$lowerleft]$\\
Rob: Yes, I won.\\
Rob: Good bye!\\
\end{tabular}
}

\caption{\label{hhinteraction} Example human-humanoid interaction while playing the game of noughts and crosses. The commands in squared brackets represent physical actions with handwriting. Example video at \url{https://youtu.be/25jdV8FN4ic}}
\end{flushleft}
\end{figure*}

Interactive humanoid robots that perceive, act, communicate and learn simultaneously are not only interesting for demonstrating robot capabilities, but they have the potential of being used to study embodied human intelligence. This paper describes a small step in these directions, where we equip a humanoid robot\footnote{\url{http://www.rethinkrobotics.com/baxter/}} with multiple input and output modalities in order to play the game of {\it noughts and crosses}, also known as `tic-tac-toe'---see Figure~\ref{hhinteraction}. These modalities allow the robot to listen to human commands, see human gazing and human drawings in the targeted game, gaze at the human player in focus, talk to human players, draw noughts/crosses, and learn from examples---all of them asynchronously. The latter capability (learning) requires very efficient forms of training to approach human-like behaviour. While previous work have studied robot learning from real interactions with the environment \cite{Levine2016,Kobber2013,Zhang2015}---mostly without verbal abilities, training an autonomous agent to learn even simple behaviours can take large amounts of experience \cite{mnih-dqn-2015}. Other previous works have addressed multimodal deep learning but in non-conversational settings \cite{WermterWEPEP04,NgiamKKNLN11,SrivastavaS14}.

While those previous works learn from raw pixels, our learning approach is semi-decoupled into two tasks: (1) learning to perceive and (2) learning to interact. It is semi-decoupled because the task of learning to interact uses multimodal perception, which requires learning from multimodal features rather than unimodal ones. Because these two tasks represent high-dimensional systems, the former uses deep supervised learning, and the latter uses deep reinforcement learning. The advantage of using this two-stage approach is that multimodal learning can be achieved more efficiently. Let us assume that our robot has to be trained to play using grids of different sizes. While the raw pixel approach would have to re-learn its behaviour for new grid sizes, our approach only re-learns to perceive and reuses its learnt behaviour to interact. Similarly, assuming that our robot has to be trained to interact using increasing amounts of verbal features (e.g. words), our approach would have to re-learn its verbal behaviour but not its vision-based perception. These two examples illustrate the benefits of our proposed approach, which can be seen as an indirect form of transfer learning. Due to the complexity of behaving in unseen environments, in this paper we use grids of one size to illustrate our multimodal deep reinforcement learning approach---see example interaction in Figure~\ref{hhinteraction}.

\section{Learning Approach for Physical Human-Humanoid Interaction} 
Our approach uses two independent but related learning tasks. First, learning to perceive in order to predict what is going on in the environment (game moves in our case). Second, learning to interact in order to decide what to do or say next. In this way, humanoid robots can learn to interact from words and game moves (a more compact set of multimodal features) rather than words or speech and pixels. Although our approach implies more efficient learning due to the more compact environment states, the latter state representation (raw multimodal features) remains to be investigated in future work.

\begin{figure}[b!]
  \begin{center}
    \includegraphics[width=1.0\textwidth]{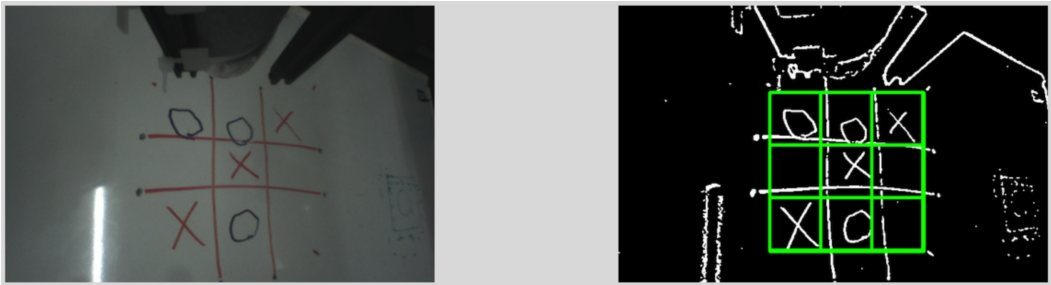}
\caption{\label{GameGrid}(Left) raw input images. (Right) Grayscale images used for game move recognition.}
  \end{center}
\end{figure}

\begin{figure*}[t!]
  \begin{center}
    \includegraphics[width=0.95\textwidth]{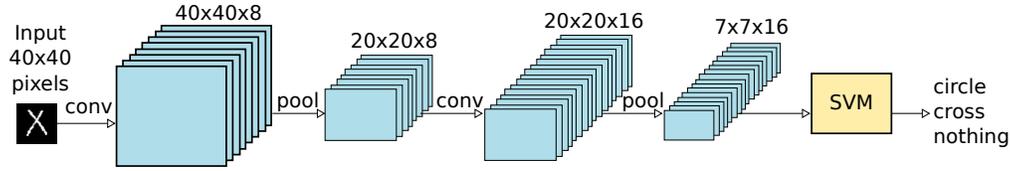}
\caption{\label{architectureDSL} Architecture of the deep supervised learner for game move recognition}
  \end{center}
\end{figure*}


\subsection{Learning to Perceive with Deep Supervised Learning} 
\label{Learning2Perceive}
We use the camera on the right arm of the robot to perceive symbols in the game grid. Rather that using a single image, we use multiple images (one per location in the grid, 9 in our case) to detect new drawings used to generate game moves. The robot continuously takes images and splits them into 9 images of 40$\times$40 pixels as shown in  Figure~\ref{GameGrid}.

\begin{figure*}[th!]
  \begin{center}
    \includegraphics[width=1\textwidth]{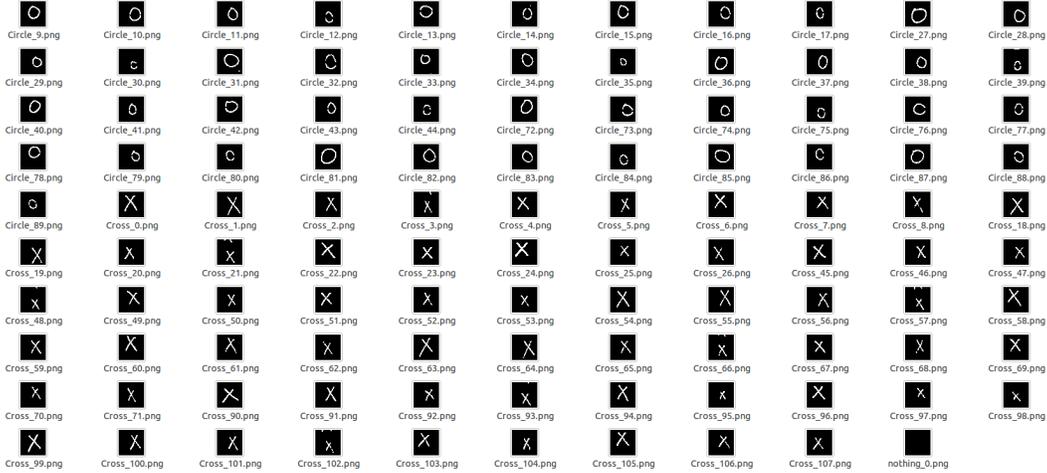}
\caption{\label{seedData} Seed training examples for training the deep supervised learner}
  \end{center}
\end{figure*}

Given a data set of the form $\mathcal{D}=\{({\bf x}_1,y_1),...,({\bf x}_N,y_N)\}$, where ${\bf x}_i$ are $n \times n$ matrices of pixel-based features and $y_j$ are class labels, the tasks is to map images to labels. In our case, the images have 40$\times$40 pixels, and the labels are \{`circle', `cross', `nothing'\}. We use a deep supervised classifier to induce function $h:{\bf X} \rightarrow Y$ out of a space of functions $H=\{h | h:{\bf X} \rightarrow Y\}$, where ${\bf X}$ are images and $Y$ are the labels. The labelling process is defined as $h({\bf x}) = \arg\max_y f({\bf x},y)$,
where $f$ is a scoring function using learnt features ${\bf x}$ derived from a Convolutional neural network \cite{LeCun2015}. To train this classifier we use a set of seed labelled images (see Figure~\ref{seedData}) to generate more images but with the drawings (circles and crosses) in randomly assigned locations. For example, an image with a cross in the middle can generate more images by shifting it to the left or right, and up or down.

Our Convolutional neural net used the architecture shown in Figure~\ref{architectureDSL} with the following layers: input layer of 40$\times$40 pixels, convolutional layer with 8 filters, RELU, pooling layer of size 2$\times$2 with stride 2, convolutional layer with 16 filters, RELU, pooling layer of size 3$\times$3 with stride 3, and the output layer used a Support Vector Machine (SVM) with 3 labels.

This classifier is used continuously often in user turns, every 100 milliseconds, to detect activity (drawings of human players) in each location of the game grid. In addition and to reduce noise, we accept a new drawing if it has been recognised at least 3 times in a row. In this way, our perception component can output game moves in the following format\footnote{The drawn symbol is inferred from who starts the game and with what symbol. For example, if the robot starts the game, we assume that the robot draws circles and the users draws crosses.}: $[who$=$usr \wedge what$=$draw \wedge where$=$middle]$.

\subsection{Learning to Interact with Deep Reinforcement Learning}
\label{Learning2Interact}
The visual perceptions above plus speech-based perceptions (words with confidence scores) are given as input to a reinforcement learning agent to induce its behaviour from interaction with the environment, where situations are mapped to actions by maximizing a long-term reward signal \cite{Sutton_Barto-1998,Szepesvari:2010}. An RL agent is typically  characterized by: (i) a finite set of states $S=\{s_i\}$; (ii) a finite set of actions $A=\{a_j\}$; (iii) a state transition function $T(s,a,s')$ that specifies the next state $s'$ given the current state $s$ and action $a$; (iv) a reward function $R(s,a,s')$ that specifies the reward given to the agent for choosing action $a$ when the environment makes a transition from state $s$ to state $s'$; and (v) a policy $\pi:S \rightarrow A$ that defines a mapping from states to actions. 
The goal of an RL agent is to find an optimal policy by maximising its cumulative discounted reward defined as 
\begin{equation}\nonumber
Q^*(s,a)=\max_\pi \mathbb{E}[r_t+\gamma r_{t+1}+\gamma^2 r_{t+1}+...|s_t=s,a_t=a,\pi],
\end{equation}
where function $Q^*$ represents the maximum sum of rewards $r_t$ discounted by factor $\gamma$ at each time step. 
While RL agents take actions with probability $Pr(a|s)$ during training, they select the best at test time, i.e. $\pi^{*}(s)$=$\arg \max_{a \in A} Q^*(s,a)$.

To induce the $Q$ function above our agent approximates $Q^*$ using a multilayer neural network as in \cite{mnih-dqn-2015}. The $Q$ function is parameterised as $Q(s,a;\theta_i)$, where $\theta_i$ are the parameters (weights) of the neural net at iteration $i$. Furthermore, training a deep RL agent requires a dataset of experiences $D=\{e_1,...e_N\}$ (also referred to as `experience replay memory'), where every experience is described as a tuple  $e_t=(s_t,a_t,r_t,s_{t+1})$. Inducing the $Q$ function consists in applying Q-learning updates over minibatches of experience $MB=\{(s,a,r,s')\sim U(D)\}$ drawn uniformly at random from the full dataset $D$. A Q-learning update at iteration $i$ is thus defined according to the loss function 
\begin{equation}\nonumber
L_i(\theta_i)=\mathbb{E}_{MB} \left[ (r+\gamma \max_{a'} Q(s',a';\overline{\theta}_i)-Q(s,a;\theta_i))^2 \right],
\end{equation}  
where $\theta_i$ are the parameters of the neural net at iteration $i$, and $\overline{\theta}_i$ are the target parameters of the neural net at iteration $i$. The latter are only updated every $C$ steps. This process is implemented in the learning algorithm {\it Deep Q-Learning with Experience Replay} described in \cite{mnih-atari-2013}. 

\begin{figure}[t!]
  \begin{center}
    \includegraphics[width=1.0\textwidth]{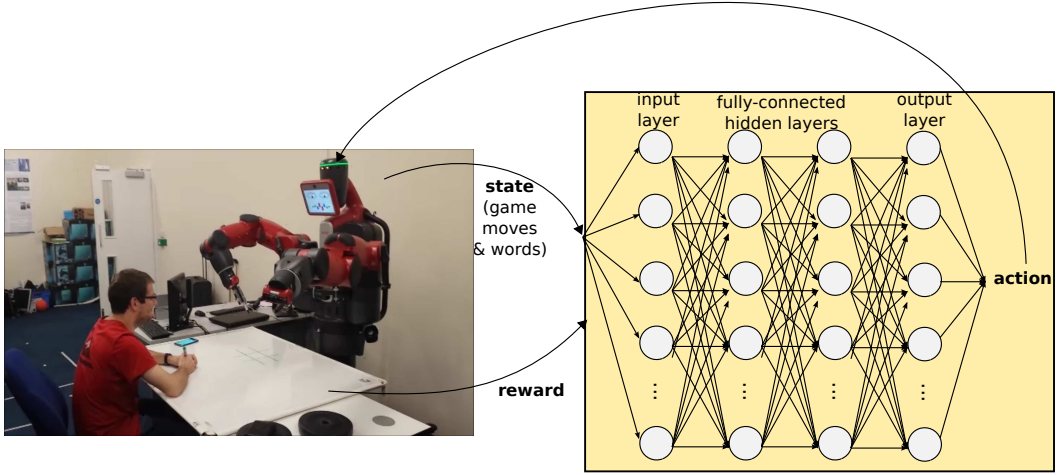}
\caption{\label{integratedSystem} (Left) Physical game-based environment, (Right) deep reinforcement learning agent for human-humanoid interaction---see text for details}
  \end{center}
\end{figure}

\paragraph{State Space}
The state space $S=\{s_i\}$ of our learning agent includes 57 features that describe the game moves and words raised in the last system and user turn. While words derived from system responses are treated as binary variables (i.e. word present or absent), the words derived from noisy user responses can be seen as continuous variables by taking confidence scores into account. Since we use a single variable per word, user features override system ones in case of overlaps. In contrast to word-based features, game moves do take into account the context or history of each game.

\paragraph{Action Space}
The action space $A$ includes 18 dialogue acts in the domain of noughts \& crosses\footnote{Actions: GameMove(gridloc=\$loc) x 9, Provide(feedback=draw), 
Provide(feedback=loose),  Provide(feedback=win), Provide(name), Reply(playGame=yes), Request(playGame), Request(userGameMove), Salutation(closing), Salutation(greeting).}. Rather than learning using all actions in every state, the action set was derived from the most likely actions,  $Pr(a|s)>0.001$, with probabilities derived from a Naive Bayes classifier trained from example dialogues. In addition, if a physical action was included in the action set, we included all valid physical actions to allow the agent explore different game moves.


\paragraph{State Transition Function}
This function is based on a numerical vector representing the last system and user word-based responses, and game history. The latter means that we kept the game move features as to describe the game state rather than resetting them at every turn. The system responses are straightforward, 0 if absent and 1 if present. The user responses correspond to the confidence level [0..1] of noisy user responses.

\paragraph{Reward Function}
It is motivated by the fact that dialogues should be human-like and game-based. It is defined as $R(s,a,s')=(BR \times w)+(DR \times (1-w))-DL$, where $BR$ is a bonus reward using the following values: 5 if the agent is about to win, 1 if it is about to draw, 0 otherwise; $w$ is a weight over the bonus reward (BR), we used $w$=0.5; $DR$ is a data-like probability of having observed action $a$ in state $s$ in the seed example dialogues; and $DL$ is used to encourage efficient interactions, we used $DL$=0.1. The $DR$ scores are derived from the same statistical classifier above, which allows us to do statistical inference over actions given states ($Pr(a|s)$). In addition, this function provided the following rewards at the end of the interaction (dialogue): 0 for loosing the game, 5 otherwise.
 
\paragraph{Model Architecture}
It consists of a fully-connected multilayer neural network with 57 nodes in the input layer  60 nodes in the first and second hidden layers, and 18 nodes (action set) in the output layer. The hidden layers use RELU (Rectified Linear Units) activation functions to normalise their weights, see \cite{NairH10} for details. Other learning parameters include the following: experience replay size=100K, discount factor=0.7, minimum epsilon=0.01, learning rate=0.001, and batch size=32.

\begin{figure*}
\begin{flushleft}
\centerline{\hspace{0.15cm} 
    \includegraphics[width=0.61\textwidth]{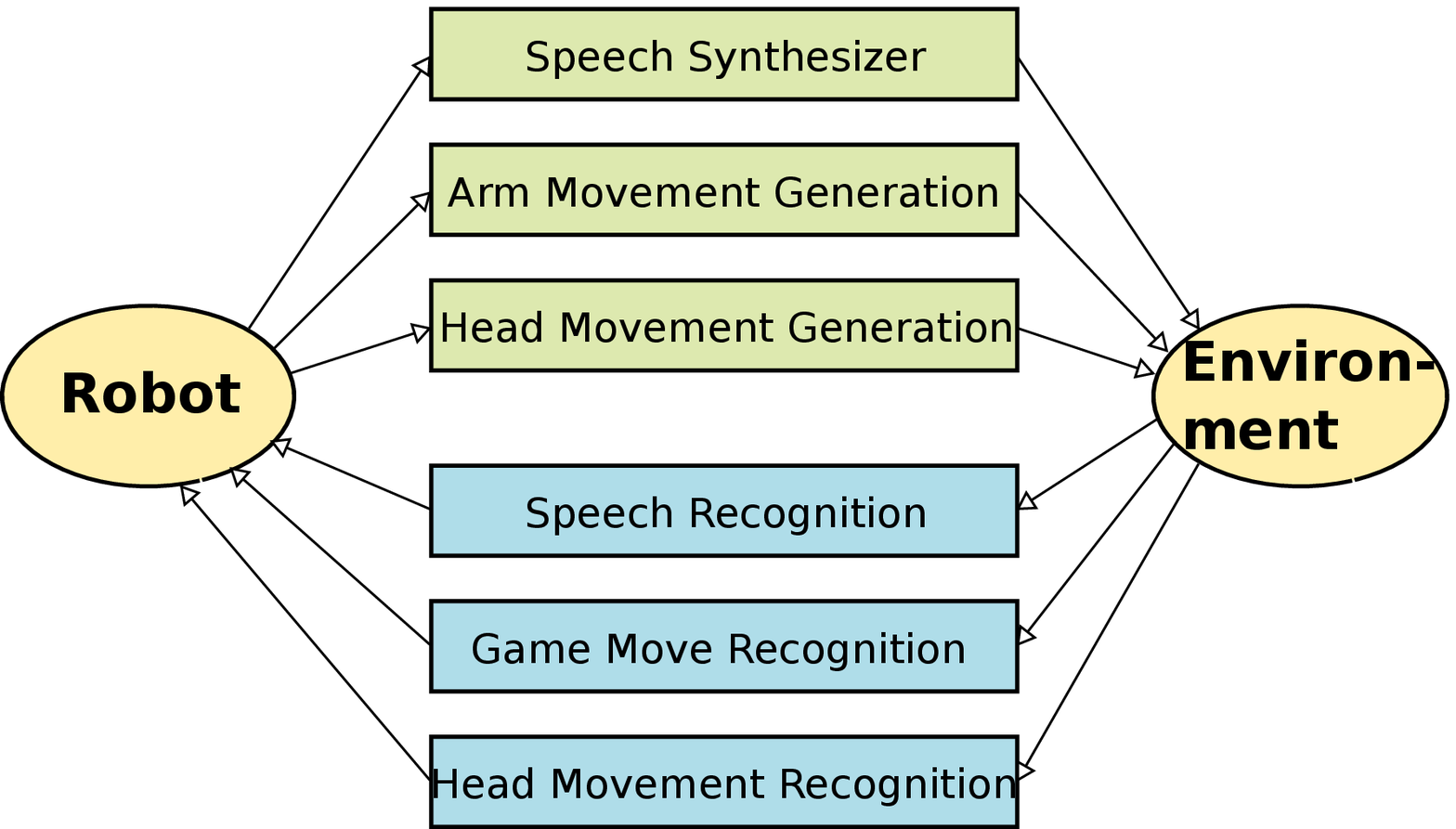}
\hspace{-0.15cm}
    \includegraphics[width=0.41\textwidth]{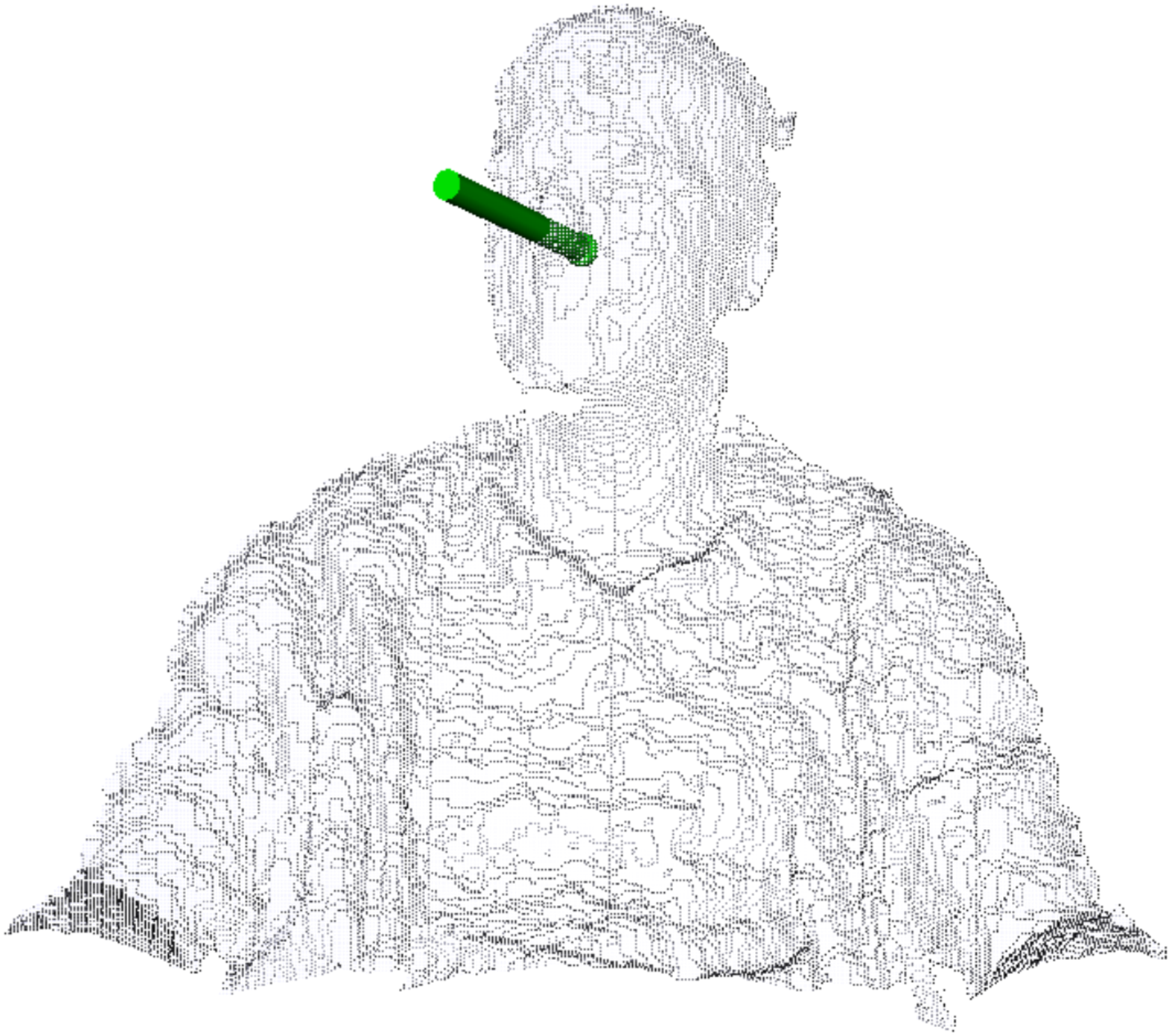}
    }
\caption{\label{integratedSystem} (Left) Components of the integrated system of the humanoid robot playing noughts and crosses, which apart from head movements that used imitation, were orchestrated by a deep reinforcement learning interaction manager. (Right) 3D head tracking is used to observe changes in orientation  based on patterns detected from the signals produced by \cite{Fanelli2013}.}
\end{flushleft}
\end{figure*}

\section{Experiments and Results} 
In this section we apply the approach and learning agent above to a humanoid robot that learns to play the game of noughts and crosses. 

\subsection{Integrated System}
Our humanoid robot was equipped with multiple modalities---including speech, touch and vision---to play the targeted game. To do that we used both off-the-shelf components and components built specifically for our ROS-based \cite{ROS} integrated system. These components run concurrently, via multi-threading, and are explained as follows.

\paragraph{Speech Recognition}
This component runs the Google Speech Recogniser on an Android App with a touch-to-speak mechanism. This component communicates the speech recognition results (also referred to as `N-best lists') to our ROS-based integrated system via Bluetooth. These speech-based perceptions are used as features in the state space of the deep reinforcement learning agent.

\paragraph{Game Move Recognition}
This component runs the vision-based perception subsystem described in ~\ref{Learning2Perceive}, see Figure\ref{GameGrid} for an illustration. Briefly, it follows the next steps: takes RGB images from a predefined initial location of the right arm (for consistent perceptions), converts them to grayscale, removes the grid, splits each image into 9 images (one image per location in the game grid), predicts the label based on an SVM classifier with learnt features (labels: circle, cross, nothing), and generates a game move based on a newly observed labels at least 3 times in a row (to avoid noise). While this component can be used to recognise all system and user game moves, it was used to recognise user game moves only. These vision-based perceptions are used as features in the state space of the deep reinforcement learning agent. 

\paragraph{Head Move Recognition}
Head tracking is used to detect changes in orientation of the human player's head (e.g. left, right, up, down, and centre). To do that we extract patterns from depth-based sensory data using a Kinect sensor and the algorithm described in \cite{Fanelli2013}---see Figure~\ref{integratedSystem} (Right). Using those patterns, we track a basic set of  movements (left, right, up, down, centre) using a threshold-based approach. This allowed the robot to know where the user is looking at in order to imitate head movements and give the impression that the robot is following the gaze of human players.

\paragraph{Speech Synthesis}
The verbalisations (in English), which correspond to translations from high-level actions derived the interaction manager to words, used a template-based approach and an off-the-shelf speech synthesizer\footnote{\url{http://mary.dfki.de/}}. The spoken verbalisations were synchronised with the face of the robot---a video played on the robot's head, which moved its eyes and mouth while speaking. Rather than using a static robot face, the video and speech started and ended simultaneously to give the impression of a synchronised talking face.

\paragraph{Arm Movement Generation}
This component receives commands from the interaction manager for drawing symbols in the game grid. Given that we assumed a static and fixed-sized game grid, the task of what and where to draw was simplified---though future work should assume dynamic game grids. While arm movements (at predefined speeds) started as soon as a command was received, it notified the interaction manager when it was executed. In this way, a future verbalisation would have to wait until the drawings were done and the arm was back at the initial position. 

\paragraph{Head Movement Generation}
This component takes the head tracking movements (from the head move recogniser) of human players as inputs in order to imitate them. Its outputs correspond to head movements to the left, right, up, down, and centre. For example, if a human player turns their head to the left and then looks at the robot, the robot does the same except that from its own spatial perspective (e.g. human turns left = robot turns right). This gives the impression that the robot is actually paying attention to the human player, and also gives more liveliness to the robot.

\begin{figure*}
\begin{flushleft}
\centerline{\hspace{0.15cm} 
    \includegraphics[scale=0.17]{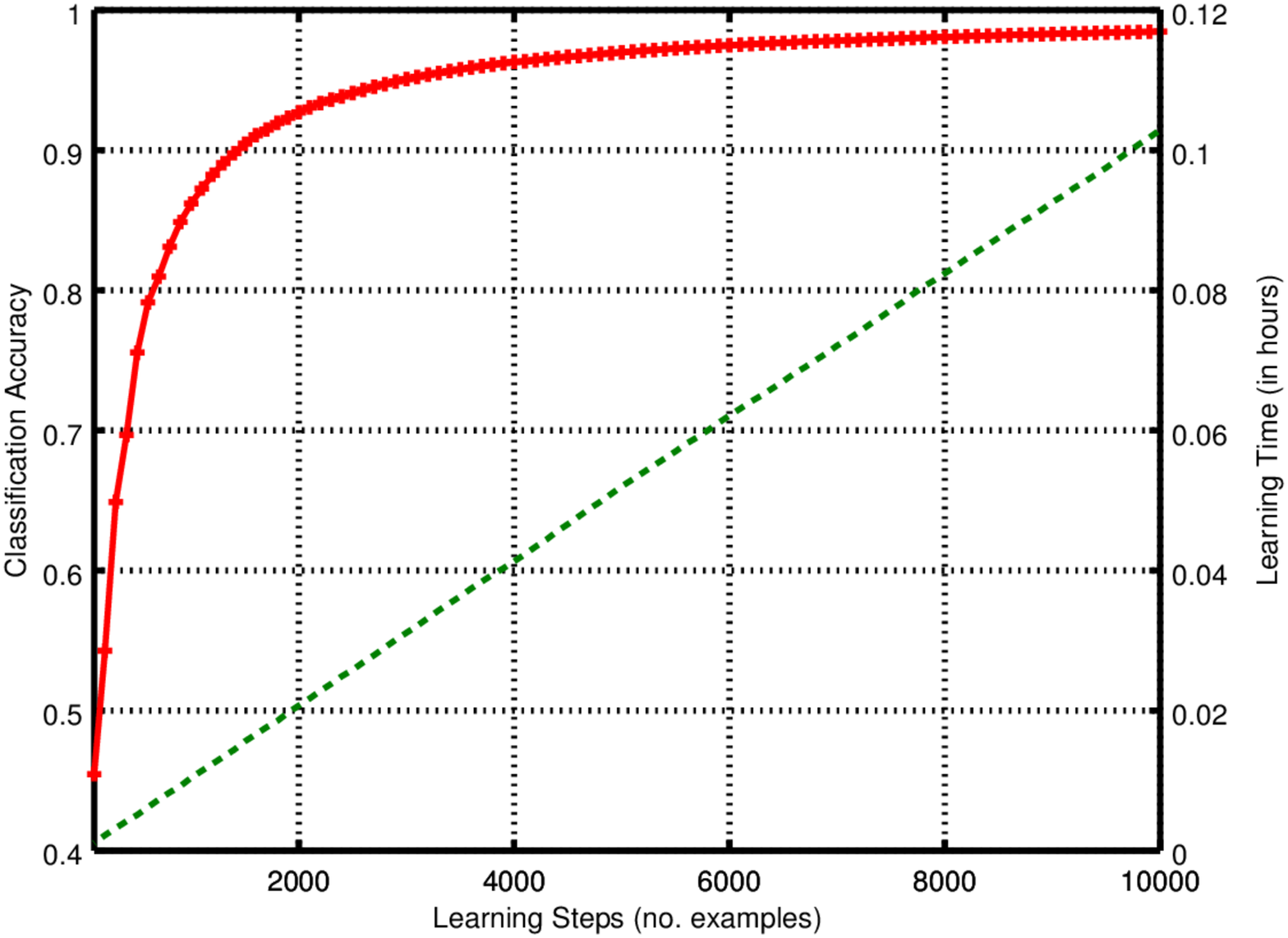}
\hspace{-0.21cm}
    \includegraphics[scale=0.17]{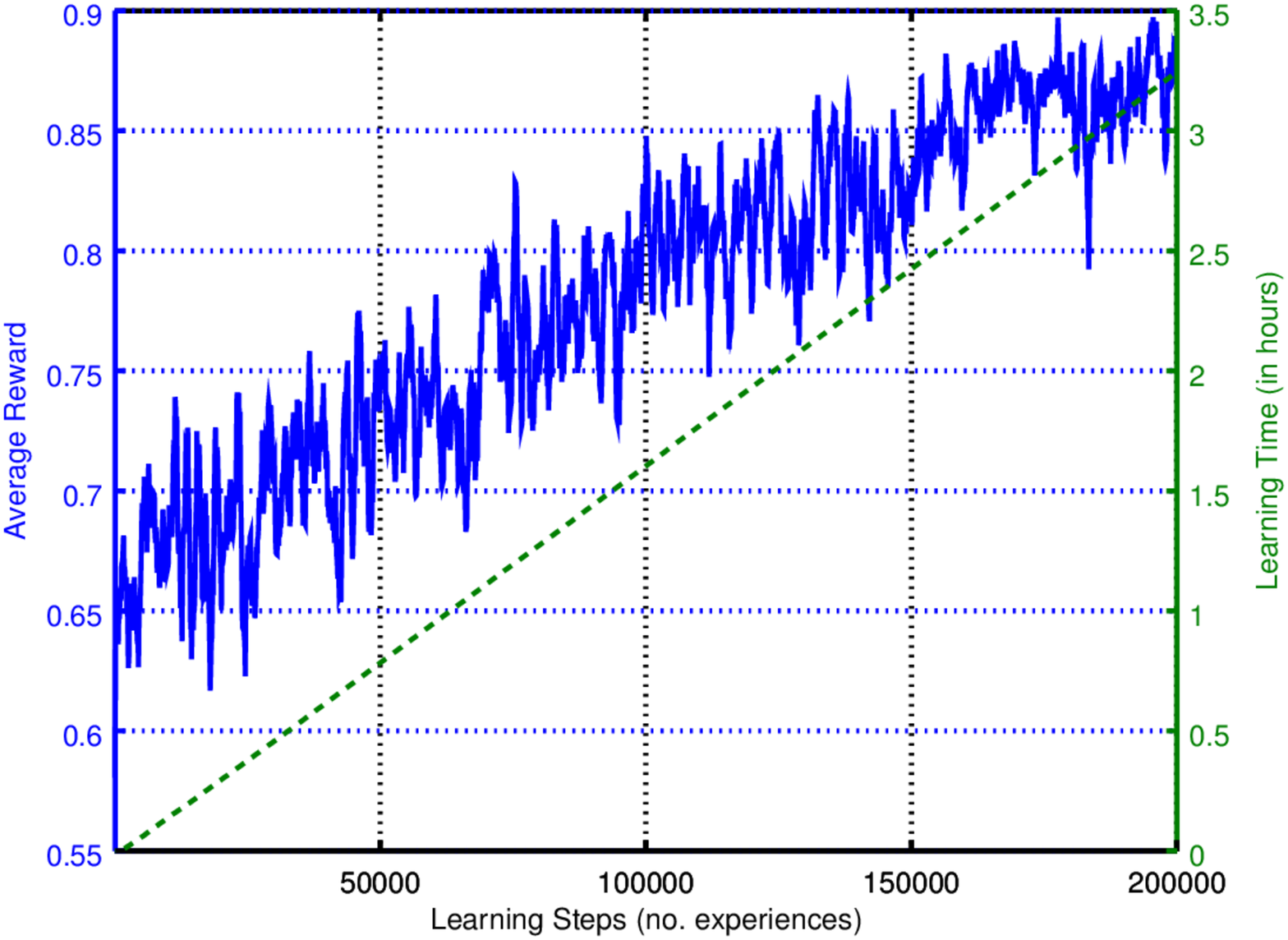} 
\hspace{-0.65cm}
    \includegraphics[scale=0.17]{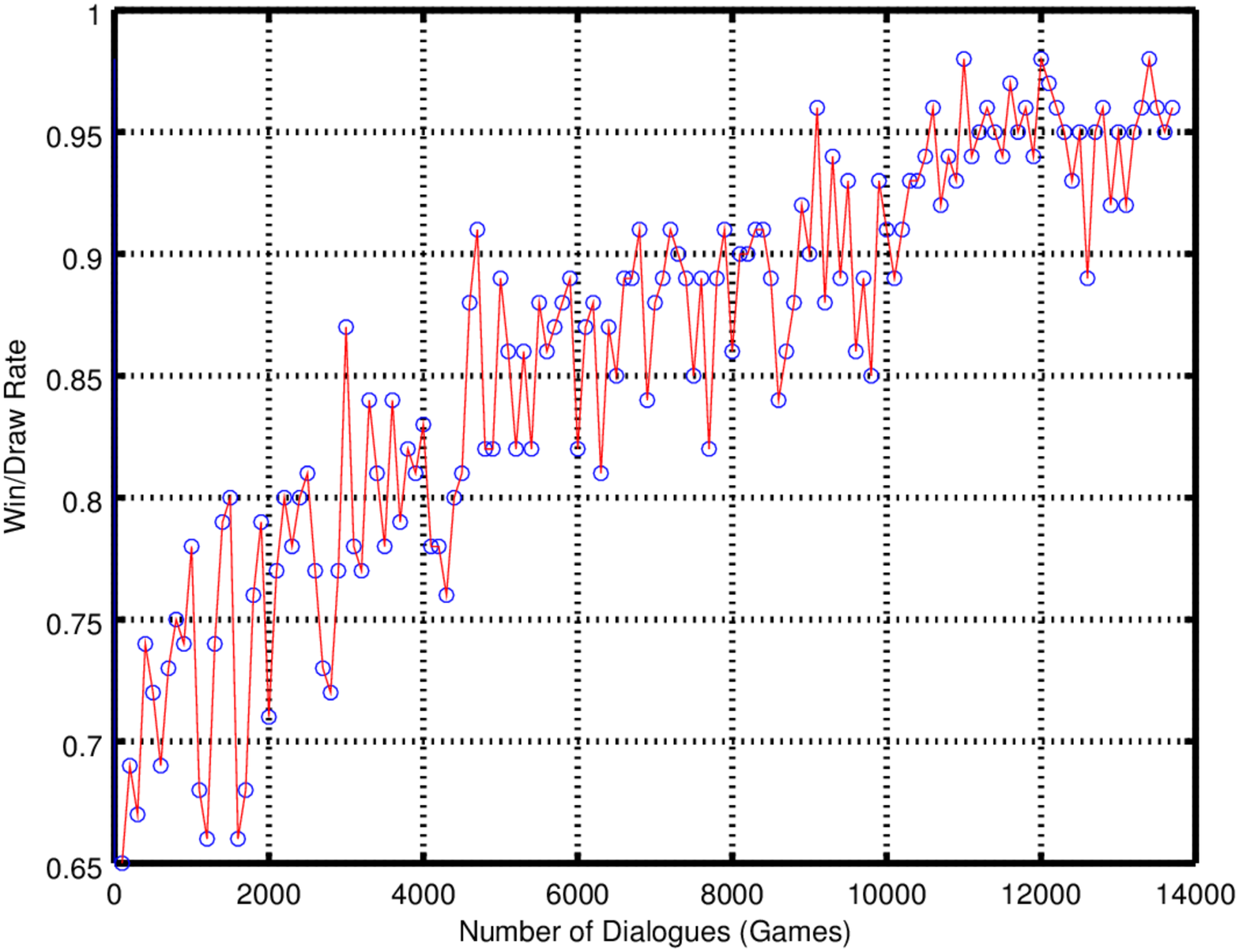}}
\caption{\label{plots} Learning curves of the deep supervised/reinforcement learners for multimodal interaction: (left) classification accuracy, (middle) average reward, and (right) win/draw rate.}
\end{flushleft}
\end{figure*}

\paragraph{Interaction Manager}
The interaction manager, based on the publicly available {\it SimpleDS} tool\footnote{\url{https://github.com/cuayahuitl/SimpleDS}} \cite{Cuayahuitl16}, orchestrates the components above by continuously receiving speech-based and vision-based perceptions from the environment, and deciding what to do next and when. Regarding what to do next, it chooses actions based on the learning agent described in Section~\ref{Learning2Interact}. While half of such actions are only verbal actions, the other half are multimodal actions. For example, communicating action $GameMove(gridloc=lowerleft)$ corresponds ``I take this one $[who$=$rob \wedge what$=$draw \wedge where$=$lowerleft]$'', where the square brackets represent a physical action (drawing a circle or cross at the given location). The policy evaluated below assumed that the robot starts playing with a default symbol=circle. Training policies with a large repertoire of multimodal actions is an interesting future research direction.


\subsection{Experimental Results} 
While the integrated system is able to produce reasonable and fluent interactions with human players (as can be observed in this  video\footnote{\url{https://youtu.be/25jdV8FN4ic}}), we focus our evaluation to learning to perceive and learning to interact from simulated interactions. 
Nonetheless, this integrated system has been used in preliminary though successful demonstrations. We tried the system with seven independent human users, and all reported successful interactions, which they enjoyed. A comprehensive evaluation with multiple human players is left as future work.


\paragraph{Deep Supervised Learner}
The task of this learner is to classify 40x40 grayscale images into three labels (circle, cross, nothing) using the SVM classifier with learnt features described in Section~\ref{Learning2Perceive}. While the classifier used a set of seed images (as shown in  Figure~\ref{seedData}), it used additionally generated images by positioning the drawings at different locations of each image. Figure~\ref{plots} (left) shows a learning curve of classification accuracy given an increasing amount of training examples, where the training time using 10K images required only 6 minutes as illustrated Figure~\ref{plots} (left, green dotted line) using a contemporary desktop computer (Core i7 with 3.4 GHz). During testing and based on 1000 randomly generated images from the seed examples, produced a classification accuracy of 99.9\%. This is an indication that our vision-based perception component was accurate enough to classify human handwriting for the targeted game. Pilot tests with human subjects reports that this component can be used to detect handwriting from different human players in a way that allow them to play the game of noughts and crosses.


\paragraph{Deep Reinforcement Learner}
The task of this learner is to select high-level multimodal actions (18 in total) based on speech-based and vision-based perceptions using the deep reinforcement learning agent described in Section~\ref{Learning2Interact}. Rather than using raw pixels, this learner takes words and game moves as features. In addition, rather than training from real human-humanoid interactions we used user simulations that provide semi-randomly generated user responses. The system actions, system responses and user responses are derived from example dialogues provided to the interaction manager. We used a set of 10 seed example dialogues as the one shown in Figure~\ref{hhinteraction}. The user simulator is semi-random because responses already chosen were not allowed. For example, drawing a symbol in the middle of the grid was only allowed if the location was empty. The goal of the agent was to induce its behaviour based on human-like behaviour (similar to the example dialogues), and to win as much as possible. 

Figure~\ref{plots} (middle) shows a learning curve of average reward according to an increasing amount of experiences, where one action is selected in each experience. A reasonable policy was found after 3 hours of training using the same desktop as the previous learner. In addition, Figure~\ref{plots} (right) shows a learning curve of win/draw rate in relation to the number of games played. These learning curves show how the proposed agent achieved successful learning from multimodal input features and multimodal actions---with no other information of the game apart from the given rewards.

\section{Related Work and Limitations}
The recent developments in machine learning, specifically in the area of deep learning, are allowing the development of more ambitious intelligent interactive systems. For example, previous interactive systems would require a substantial amount of effort in feature selection. Now, deep supervised learners can be trained from learnt features \cite{LeCun2015}, and deep reinforcement learners can be trained to induce their features and policy jointly \cite{mnih-dqn-2015,MnihBMGLHSK16}. Despite of these advances, applying deep learning to interactive robots is far from trivial. For example, the robot described in \cite{Zhang2015} was trained to carry out target reaching from pixels, which was successful in simulation but failed when tested in the real environment. The behaviours of other robots have been induced with reasonable success---though they usually do not target multiple modalities as in this paper. In addition, previous works teaching agents to play noughts and crosses assume perfectly drawn grids and symbols without any multimodal inputs and outputs \cite{Boyan92modularneural,Siegel2001,SteegDW15}. Our robot assumes perception from imperfect human drawings, speech-based responses, and gestures. Training robots to perceive, act and communicate using multiple modalities is important to bring them to end users \cite{Mavridis2015ras,HC2015aisb}. Our humanoid robot has, to our knowledge, the first system that learns to perceive, act and communicate using deep (reinforcement) learning. This system can be considered as data efficient because it used only 108 seed images and 10 seed dialogues to bootstrap the learning environment and agent.

Although our robot system is reasonably advanced, it has a number of limitations that encourage interesting research avenues. First, our robot assumes a fixed-size grid, and grids of different sizes and shapes (as drawn by humans) remain to be explored. This would represent a major upgrade because this direction addresses joint perception and gestures (arm movements). The former requires more robust perception due to unknown drawings, and the latter requires the robot to draw symbols in non-predefined locations and of different sizes. 
Second, our robot uses push-to-talk speech-based interaction. It would be interesting to compare this approach with a microphone array so human players can have more natural interactions.
Third, our robot used a small vocabulary (about 40 different words) and larger vocabularies remain to be explored as well, specially if we would like a chatty robot rather than a robot that plays the game using the same verbalisations over and over again. 
Fourth and in a similar vein, our robot uses predefined templates for language generation, and including a situated language generator (e.g. \cite{DethlefsC15}) in the training process would contribute towards more natural interactions.   
Fifth, the robot learns its behaviour offline and uses its best behaviours while playing with human users (without further learning). It would be interesting to incorporate other forms of learning to train or retrain the robot as it collects data from real interactions \cite{Argall2009,Kobber2013}. Unsupervised, semi-supervised and active learning could be useful here to learn from unlabelled examples or labelled ones but in a more efficient way than standard supervised learning. Sixth, it would be interesting to integrate more complex versions of noughts and crosses and other games using deep learning \cite{CuayahuitlKL15}, which represents a niche for investigating more efficient learning. The more complex the human-humanoid interactions the more features and actions, which will be reflected in slow or infeasible learning. Last but not least, evaluations with end users remain to be carried out, specially with unknown human players in public spaces.

\section{Concluding Remarks}
Robot systems that interact with their environment by perceiving, acting, communicating, and learning often face a challenge in how to bring these different concepts together. We describe a general approach for training a robot to interact with the world using multiple modalities. Rather than training the robot directly from raw pixels only, the proposed approach simplifies the overall
learning task into two stages: learning to perceive and learning to interact. We tested our approach by training the Baxter humanoid robot to play the game of nought and crosses. Our experimental
results using simulations report that learning to perceive achieved 99.9\% of classification accuracy, and learning to interact achieved a win/draw rate of 98\%. A pilot test reported reasonable interactions using the proposed approach in a multimodal integrated system. In addition, our system showed to be data-efficient due to the amount of data used to induce the simulated environment (108 seed images and 10 seed dialogues), which is relevant for trainable robot systems from example demonstrations with end users—as pointed out by \cite{Vollmer2016}. This is the first deep (reinforcement) learning system that learns to perceive, act and communicate. Although functional for demonstration purposes, it requires a number of improvements before it can be released in the wild. The previous section outlines exiting future directions in interactive intelligent humanoids.


\footnotesize
\bibliographystyle{abbrv}
\bibliography{ibaxter}

\end{document}